**Near-perfect photo-ID of the Hula painted frog with zero-shot deep local-feature matching**

Maayan Yesharim[1], R. G. Bina Perl[2], Uri Roll[3], Sarig Gafny[4], Eli Geffen[1], Yoav Ram[1,5,*]

1. School of Zoology, Faculty of Life Sciences, Tel Aviv University, Tel Aviv, Israel
2. Department of Terrestrial Zoology, Senckenberg Research Institute and Natural History Museum, Frankfurt, Germany
3. Mitrani Department of Desert Ecology, The Jacob Blaustein Institutes for Desert Research, Ben-Gurion University of the Negev, Midreshet Ben-Gurion, Israel
4. Faculty of Marine Sciences, Ruppin Academic Center, Michmoret, Israel
5. Safra Center for Bioinformatics, Tel Aviv University, Tel Aviv, Israel

* Corresponding author: yoavram@tauex.tau.ac.il

13.1.2026

**Abstract**

Accurate individual identification is essential for monitoring rare amphibians, yet invasive marking is often unsuitable for critically endangered species. We evaluate state-of-the-art computer-vision methods for photographic re-identification of the Hula painted frog (*Latonia nigriventer*) using 1,233 ventral images from 191 individuals collected during 2013–2020 capture–recapture surveys. We compare deep local-feature matching in a zero-shot setting with deep global-feature embedding models. The local-feature pipeline achieves 98% top-1 closed-set identification accuracy, outperforming all global-feature models; fine-tuning improves the best global-feature model to 60% top-1 (91% top-10) but remains below local matching. To combine scalability with accuracy, we implement a two-stage workflow in which a fine-tuned global-feature model retrieves a short candidate list that is re-ranked by local-feature matching, reducing end-to-end runtime from 6.5–7.8 hours to ~38 minutes while maintaining ~96% top-1 closed-set accuracy on the labeled dataset. Separation of match scores between same- and different-individual pairs supports thresholding for open-set identification, enabling practical handling of novel individuals. We deploy this pipeline as a web application for routine field use, providing rapid, standardized, non-invasive identification to support conservation monitoring and capture–recapture analyses. Overall, in this species, zero-shot deep local-feature matching outperformed global-feature embedding and provides a strong default for photo-identification.

**Introduction**

Amphibians are the most threatened vertebrate class: 40.7% of species are listed by the IUCN Red List as Vulnerable, Endangered, or Critically Endangered (Luedtke et al. 2023). The Hula painted frog (*Latonia nigriventer;* henceforth 'Latonia') is among the world's rarest amphibians and the only extant member of the genus Latonia (Perl et al. 2017). Its population size is estimated at ~240 breeding adults and an estimated effective population size of 17–36 at its main known breeding site in northern Israel (Perl, Geffen, et al. 2018). Monitoring this species' population is critical to enable its long-term persistence. Mark–recapture methods can help track its population size and trends (Ferner 2007), as well as the effectiveness of conservation measures. Nevethreless, traditional marking methods such as toe-clipping or passive integrated transponder (PIT) tagging are impractical for this at-risk species as they are invasive and raise welfare and detection concerns (Nemesházi et al. 2025; *Measuring and Monitoring Biological Diversity* 1994).

Individual re-identification using photographs is now widely used for non-invasive capture-recapture studies (Bolger et al. 2012; Matthé et al. 2017). Latonias have individually distinctive ventral spot patterns that have been proven to be a useful tool for individual re-identification . However, manual matching of ventral photographs is becoming time-consuming and error-prone as the image database grows over multiple surveys, a limitation already recognized in other photographic capture-recapture systems (Matthé et al. 2017; Perl, Gafny, et al. 2018). Consequently, an AI-based re-identification system can accelerate image matching and improve identification accuracy, reduce analyst workload and enable faster, standardized processing of large photo datasets. Similar automated computer-vision pipelines are increasingly used in other wildlife systems (de Lorm et al. 2023). Recent automated re-identification approaches broadly fall into (i) global-feature embedding models, which represent each image with a single feature vector for similarity-based retrieval, and (ii) local-feature matching pipelines, which represent each image by sets of local features matched across images. Hybrid systems that combine global and local features are used to balance accuracy and computational cost (Čermák et al. 2024; Nepovinnykh et al. 2024).

Here, we evaluate state-of-the-art computer-vision approaches for individual re-identification of the endangered Hula painted frogs. We assess computer-vision approaches for individual re-identification of the endangered Hula painted frog (*Latonia nigriventer*), comparing pretrained global-feature embedding models and local-feature matching pipeline. Using a manually labeled dataset of 1,233 photographs from 191 individuals, we show that a zero-shot local-feature pipeline achieves 97.8% top-1 closed-set identification accuracy. Pretrained global-feature models perform poorly without training and, although fine-tuning markedly improves their accuracy, they still underperform relative to local-feature matching, underscoring an accuracy–computation trade-off between fast global-feature retrieval and slower local-feature matching.

Building on these results, we implemented a two-stage re-identification system. In Stage 1, a fine-tuned global-feature model embeds a query image and retrieves a ranked candidate list from the gallery. In Stage 2, we apply local-feature matching only to the top candidates (e.g., top-100) to re-rank and select the final match, combining the speed of global retrieval with the accuracy of local matching. We developed a web application around this pipeline and deployed it for routine field monitoring. The application allows field researchers to upload images, review ranked matches with scores and gallery comparisons, and confirm or reject re-identifications, providing a practical and scalable tool for the long-term, non-invasive, accurate monitoring of the critically endangered Hula painted frog.

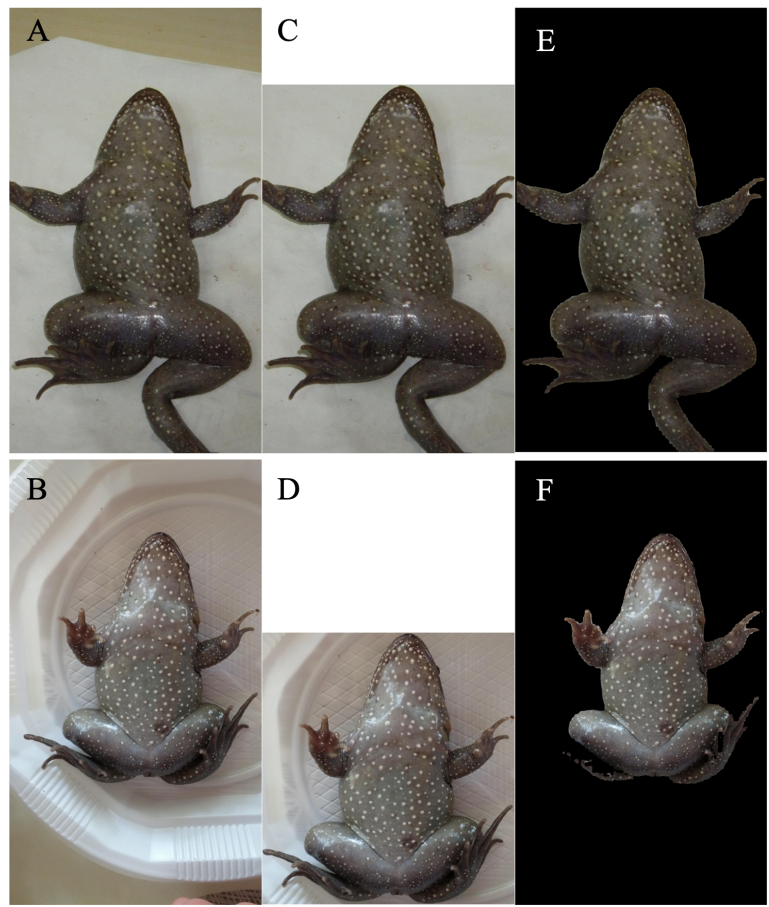

**Figure 1. Preprocessing of Hula painted frog images. (A,B)** Original images. **(C,D)** Images cropped to bounding box by MegaDetector (Beery et al. 2019). **(E,F)** Images masked with Segment Anything (Kirillov et al. 2023). The two rows show the same individual (43) from two different sessions in 2015; top row: 2020-5/229/IMGP0759.jpg; bottom row: 2015-5b/43/20150601_125108.jpg.

**Methods**

**Data.** The study draws on a dataset comprising a total of 1,233 color JPEG images of *L. nigriventer* collected during capture–recapture surveys between Dec 2013 and August 2020. Each frog was photographed with a small waterproof camera (Pentax WG-II) or a handheld digital SLR camera under ambient daylight and at a distance of 40–60 cm; multiple photos were taken per individual to capture minor pose and focus variation representative of real daytime monitoring conditions. The photos focus on the ventral view of the animals due to the abundance of white spots that can be used to distinguish between individuals (Figure 1).

We (RGBP) labeled the dataset using Wild-ID (Bolger et al. 2012) to screen for matching frog images from previous days and then confirming the match by eye. We assigned unique labels to 191 frog individuals, each of them with between 1 and 20 images (median 5). Of the 191 identified frogs, 137 were only identified once, 43 were identified twice, eight were identified three times, two were identified four times, and one frog was identified five times. For each sampling date, there are between 1-18 images per frog individual (median 4; 90% CI 2 to 9 images).

When finetuning MiewID (see below), we split the labeled dataset into class-disjoint training and validation sets. Identities were first grouped by how many distinct sampling dates they appeared on. Within each group we randomly assigned ~20% of identities to the validation set and kept the rest for training, so the distribution of temporal recaptures was maintained. All images of a given identity were kept on the same side of the split to prevent leakage. This procedure yielded a training set with 1,001 images from 152 labels and a validation set with 232 images from 39 labels, which were used exclusively for model evaluation and threshold calibration.

**Local-feature models**

**ALIKED + LightGlue.** The ALIKED model is a deep network (0.67M parameters) that, given an image, jointly detects keypoints (2D pixel coordinates indicating the locations of distinctive local features) and extracts geometrically robust local descriptors, or features, via a convolutional neural network (CNN) with a sparse deformable descriptor head (Zhao et al. 2023). We used ALIKED with 128-dimensional features, maximum number of keypoints 1432, and detection threshold 0.01.

The extracted features are then matched across image pairs with the LightGlue model (13.7M parameters), a transformer-based local feature matcher (Lindenberger et al. 2023). From the sets of keypoints and descriptors of a pair of images, LightGlue uses self-attention and cross-attention to predict a set of matches and their scores, i.e., keypoint *i* in image 1 matches keypoint *j* in image 2 with confidence score $c_{ij}$. The number of keypoint matches is the match similarity, a score between 0 (most disimilar) and the maximum number of features (most similar).

ALIKED was previously pretrained on standard feature-matching benchmarks with ground-truth geometric supervision (e.g., HPatches/MegaDepth-style data), rather than a single named dataset. LightGlue was previously pretrained on synthetic homographies of real images and on 170k images from the Oxford-Pairs 1M distractors dataset using ALIKED local features. Notably, neither model was pretrained on wildlife or re-identification tasks.

We chose ALIKED because its lightweight architecture with a deformable descriptor head offers an attractive trade-off between matching accuracy and computational efficiency compared to heavier local feature extractors. We chose LightGlue because it is more efficient than previous deep geometric matcher, e.g., SuperGlue, while maintaining high accuracy. The models are publicly avaliable through GitHub (https://github.com/cvg/LightGlue).

**Classical SIFT-based baseline.** As a classical local-feature baseline, we used SIFT keypoints and descriptors followed by brute-force matching and geometric verification (Lowe 2004; Fischler and Bolles 1981). For each image, keypoints and 128-dimensional descriptors were extracted using the OpenCV implementation of SIFT (Bradski 2000). For every image pair, we performed all-to-all descriptor matching with a brute-force L2 matcher and applied Lowe's ratio test to retain only unambiguous correspondences between the two nearest neighbors in descriptor space. The retained matches were then filtered using RANSAC homography estimation, and the match similarity between two images was defined as the number of matches after RANSAC filtering.

**Preprocessing**. We manually rotated the image so that the head of the frog is at the top. We used *Segment Anything* (Kirillov et al. 2023) to outline the frog in each photo by giving the model the coordinates of the image center as its starting point. The predicted outline was then used to mask the background with black (Figure 1).

**Global-feature models**

**MiewID**. The MiewID-msv3 model encodes 440×440 pixel RGB images to a 2,152-dimensional embedding vector (Otarashvili et al. 2024). MiewID started from a convolutional network, EfficientNetV2-RW-M (51M parameters), pretrained on ImageNet-1K and optimized for training speed and parameter efficiency (Tan and Le 2021). MiewID-msv3 was finetuned for animal re-identification using 64 wildlife species datasets (Otarashvili et al. 2024), and these finetuned weights are loaded during model initialization. The original fine-tuning employed a sub-center ArcFace classifier head and loss function (k=3, scale 51.5, dynamic margin started at 0.5) (Deng et al. 2020). The model is publicly available through HuggingFace (https://huggingface.co/conservationxlabs/miewid-msv3).

**Fine-tuning MiewID with Hula painted frog images.** We finetuned MiewID using only the training set. We initialized a new ArcFace head, which maintains a class weight vector for every identity, and trained it from scratch. Images were augmented with random resized crop, color jitter, random grayscale, and gaussian blur (random erasing did not improve performance). All model layers were finetuned except for BatchNormalization layers, which remained frozen. We used the AdamW optimizer (Loshchilov and Hutter 2019) with learning rates 2e-5 for the backbone and 1e-3 for the ArcFace head and weight decay 1e-4. Learning rates decayed to 1e-6 by a cosine schedule. Training was conducted with batch size 24 on a workstation using NVIDIA RTX A4000 GPU with 16GB. Validation was run every five epochs, recording top-1 accuracy and mAP@R (mean average precision at R, shown to exhibit lower training-time noise compared to other retrieval metrics (Musgrave et al. 2020)). We controlled per-batch class diversity with an m-per-class sampler by testing m=1 and m=4, so each update saw either 24 or 6 distinct identities, respectively. For the ArcFace loss we compared 1 versus 3 sub-centers per class (k=1 and k=3) and explored four scale values (s=30, 40, 48, or 64) and four angular margins ($\alpha$=0.30, 0.40, 0.45, and 0.50), following the ArcFace formulation (Deng et al. 2022) and its sub-center extension (Deng et al. 2020). All model variants were trained for up to 100 epochs. Runs were terminated early if mAP@R remained below 0.2 at epoch 30 or 0.22 at epoch 60.

**MegaDescriptor.** The MegaDescriptor model has several variants. For example, the L-224 model variant encodes 224×224 pixel RGB images to a 1,536-dimensional embedding vector (Čermák et al. 2023). MegaDescriptor-L-224 is a Swin-Transformer (Swin-L, patch 4, window 12, 228.8M parameters) pretrained on ImageNet-22K(+1K) (Liu et al. 2021). MegaDescriptor models were finetuned for animal re-identification using 29 publicly available datasets provided through the WildlifeDataset toolkit (Čermák et al. 2023), and these finetuned weights are loaded during model initialization. Fine-tuning employed an ArcFace loss (Deng et al. 2022) (margin 0.5, scale 64). The MegaDescriptor models are publicly available through HuggingFace (https://huggingface.co/collections/BVRA/megadescriptor).

**Preprocessing**. We manually rotated the image so that the head of the frog is at the top. For each image, we first cropped it to a bounding box around the frog using a zero-shot prediction from MegaDetector (Beery et al. 2019). We then zoomed in x2 on the center, cropped to a square, and resized to 440×440 (MiewID) or 224x224 (MegaDescriptor-L-224) pixels. Finally, we normalized the images following the procedure described in the model card.

**Cosine similarity.** Identification is performed entirely within the global features embedding space. Images are first embedded using the model to a high-dimensional vector (2,152 or 1,536 dimensions for MiewID and MegaDetector-L-224, respectively). The embedding vectors are then $\ell^2$-normalised to lie on a unit hyperball. Pairwise cosine (inner-product) similarities of embedding vectors are then computed in a single matrix multiplication. Cosine similarity is between -1 (most disimilar) and 1 (exact match).

**Two-stage pipeline.** Given a set of query images of a single frog and a gallery of images of previously photographed frogs, we adopt a two-stage retrieval pipeline. First, we use MiewID to compute global embeddings and rank all gallery images by their similarity to each query image. For each query, we then retain the top-k candidates (with k=100 in our experiments) and apply ALIKED+LightGlue to re-rank these candidates based on their local match similarity. Finally, we pool the re-ranked candidates from all query images and select the gallery image with the highest match similarity as the predicted identity of the focal frog.

**Evalutation.** To evaluate the models, we compute a similarity score for every pair of images (cosine similarity for MiewID and MegaDescriptor; match similarity for ALIKED+LightGlue). Image pairs taken on the same date are discarded because they are trivially either same- or different-individual comparisons. Each remaining image is then treated in turn as a query, with all images from different dates serving as references, after removing individuals photographed on only a single date; thus, the task is a closed-set identification problem. For each query, we rank the reference images by their similarity score and compute top-k accuracy as the probability that the true identity appears among the k highest-ranked references. We report top-k accuracy both at the image level and at the identity level.

For running time, we used Python's *time.perf_counter* at the beginning and end of the evaluation script. Thus, our running time includes time to load the models from checkpoints, to load the data, and to print a few messages, but not the time to load the Python interperter or import modules like NumPy or PyTorch.

## Results

**Global-feature models.** We started by evaluating the two global-feature models, MiewID and MegaDescriptor, which were previously pretrained on wildlife re-identification datasets (Čermák et al. 2023; Otarashvili et al. 2024). These datasets are dominated by mammals, with only a few non-mammalian species (turtles and sharks) for which the models showed large variation in performance. This suggests possible limitations in generalization to non-mammal species, and specifically, to ventral images of Latonia.

In a zero-shot setting, that is, without finetuning the models on Hula painted frog images, both global-feature models produced modest results at best (Table 1): evaluated on the entire labeled set,

MiewID obtains top-1 identity accuracy of 8.7% and top-10 closed-set identification accuracy of 36.2%; MegaDescriptor obtained even lower values. Consequently, we concluded that these models require fine-tuning before they can be effectively used with Hula pained frog images. We decided to proceed with MiewID because it outperformed MegaDescriptor in our evalutation and in previous evaluations (Otarashvili et al. 2024). It is also a smaller (51M vs 228M parameters) convolutional (rather than ViT) model, making it easier to fine-tune as it requires less compute and memory.

We finetuned the MiewID model on our training set (1,001 images from 152 identities) and evaluated it on the validation set (232 images from 39 identities). Across 34 hyper-parameter configurations, the best result yielded mAP@R = 0.325, obtained with a single image per identity per batch (m=1), single ArcFace center per identity (k=1), scale, s=30, and angular margin, α=0.4, demonstrating that maximizing identity diversity within each batch and using a lower scale and softer margin—choices consistent with small-batch theory from the ArcFace literature—yielded the greatest retrieval accuracy, while additional sub-centers or duplicate positives confered no benefit. Increasing the number of sub-centers, k=3, or the number of images per identiy per batch, m=4, systematically depressed mAP@R by up to 0.08, indicating that with only ~7 images per individual neither extra prototypes nor repeated positives improve class separation. Collectively these results support adopting the streamlined configuration m=1, k=1, s=30-40, α=0.35-0.40, consistent with similar findings by others (Berton et al. 2025).

Following our supervised fine-tuning, MiewID improved on the validation set: top-1 identity accuracy increased from 19.4% to 60.2% and top-10 identity accuracy increased from 64.1% to 91.3% (Table 1). Hence, we characterize the zero-shot performance as mediocre for the Hula painted frog dataset and the finetuning step as essential for individual recognition with global features.

**Local-feature models.** Next, we evaluated the local-feature pipeline using ALIKED to extract keypoints and descriptors, and LightGlue to match the keypoints. ALIKED+LightGlue performed extremely well, obtaining a near-perfect 97.8% top-1 closed-set identification across the labeled dataset. For comparison (Table 1), we evaluated classical local-feature baselines on the same task. We used SIFT as the classical extractor , and brute-force matching, Lowe's ratio test, and RANSAC as the classical matcher (Fischler and Bolles 1981; Lowe 2004). The ALIKED+LightGlue combination perfromed best (top-1 identity accuracy 97.8%), clearly topping ALIKED+classical (86.7%), SIFT+LightGlue (81%), and SIFT+classical (45%).

**Open-set matching.** To evaluate whether the match similarity (number of ALIKED+LightGlue matches) can be used to reject incorrect top-1 candidates in an open-set re-identification setting, we plotted histograms of the number of matches for image pairs of different individuals and for pairs of the same individual photographed on different dates (Figure 4). Pairs of different individuals (blue) are concentrated at low similarity scores, whereas pairs of the same individual (orange) cluster at much higher scores, with only a limited region of overlap. We also show examples of a clearly different pair (Figure 2A), a high-scoring but incorrect match (Figure 2B), and a clearly correct match (Figure 2C), and how the corresponding match similarities fall within these histograms (Figure 4). This suggests that a single threshold on match similarity can remove most false positives while discarding relatively few true matches.

To find such a threshold, we plotted a precision–recall curve (Figure 4). For all pairs of images taken on different dates, we treated pairs of the same individual as positives and pairs of different individuals as negative and varied the similarity threshold to classify the top-1 candidate as either accepted or rejected. The resulting curve remains close to the upper-right corner, indicating that high precision can be achieved at relatively high recall. In particular, a threshold chosen to obtain recall 0.95 still yields precision 0.78, whereas a threshold chosen to obrain precision 0.95 yields recall 0.85, illustrating that the match similarity provides a useful control over the trade-off between missing true matches and rejecting false positives.

**Running time.** We found that LightGlue is computationally expensive: both SIFT+LightGlue and ALIKED+LightGlue ran for 6.5-7.8 hours on a single 16GB GPU on the entire labeled dataset, whereas SIFT+classical and ALIKED+classical were considerably less expensive and took only 2.5 hours and 1.15 hours, respectively (ALIKED uses a GPU whereas SIFT and the classical matcher run on a CPU). However, in terms of speed, the global-feature models (MiewID and MegaDescriptor) were much faster, taking less than a minute to process the entire dataset. This is due to (1) the cosine similarity being fast to compute by matrix multiplication of the embeddings, followed by normalization, and (2) the global feature models being able to work on batches of images (we used batch size of 32).

**Two-stage pipeline.** Therefore, we tested a two-stage pipeline to reduce running time without compromising accuracy. In this pipeline, we first apply the finetuned MiewID to find the top-100 references per query, and then apply ALIKED+LightGlue to re-rank the candidates. This provided a good compromise: top-1 identity accuracy on the validation set (rather than the full set, as we finetuned MiewID on the train set) went from 100% with ALIKED+LightGlue to 98.1% with the two-stage pipeline, while runing time was reduced from 7-8 hours to 38 minutes on the entire labeled dataset (Figure 3A).

**Number of keypoints.** One way to reduce the running time is to decrease the maximum number of keypoints, M, produced by ALIKED so that LightGlue has fewer keypoints to match (one can also tweak other hyper-parameters (Lindenberger et al. 2023)). In our main analysis, we used $M$=1432. Decreasing $M$ to 600-800 reduces runtime substantially while keeping identification accuracy high over a broad range (Figure 3B). Interestingly, top-1 peaks at $M$=800 (98.5%) and is slightly lower at $M$=1432 (97.8%), while higher-order top-k accuracy changes little; this is plausible when candidates are ranked by the number of matches, since adding many lower-quality keypoints can introduce additional accidental matches to similar-looking candidates and occasionally swap the top-ranked identity without affecting whether the correct identity appears among the top few. Going forward, we keep $M=1432$ as the default for consistency across experiments, and treat $M=800$ as a computationally cheaper alternative when runtime is a limiting factor (with no observed accuracy loss on this dataset).

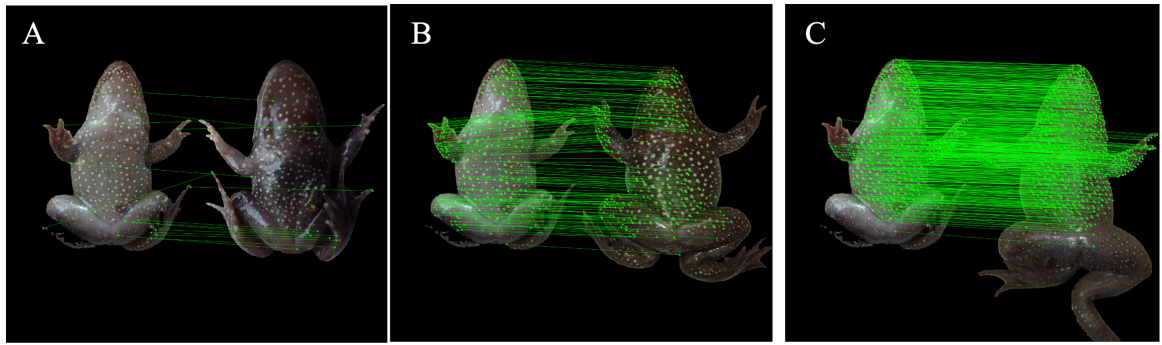

**Figure 2. Local feature matching of Hula painted frog individuals.** Examples of deep local feature matching between different individuals (A, 15 matches; B, 189 matches) and the same individual photographed on different dates (C, 666 matches). Green lines connect matching keypoints.

**Table 1. Model evaluation on Hula painted frog images.** Models evaluated on the labeled set with 1,233 images, including 542 images from 54 identities that appear on at least two different dates except for MiewID-FT, which was finetuned on the train set and evaluated on the validation set with 233 images, including 103 images from 11 identities that appear on at least two different dates.

| *Feature extractor* | Matcher | Running time (sec) | Top-1 ID accuracy | Top-3 ID accuracy | Top-10 ID accuracy | Top-1 accuracy | Top-3 accuracy | Top-50 accuracy | Top-100 accuracy |
|---|---|---|---|---|---|---|---|---|---|
| *ALIKED* | LightGlue | 28,237 | 97.8% | 98.7% | 99.3% | 97.8% | 98.7% | 99.4% | 99.6% |
| *ALIKED* | Classical | 4,368 | 86.7% | 92.4% | 94.6% | 86.7% | 92.1% | 96.7% | 97.4% |
| *SIFT* | LightGlue | 24,704 | 81.0% | 85.2% | 89.3% | 81.0% | 84.7% | 92.4% | 93.9% |
| *SIFT* | Classical | 8,410 | 45.0% | 56.6% | 68.1% | 45.0% | 56.5% | 79.3% | 86.3% |
| *MiewID-msv3* | Cosine similarity | 47.7 | 8.7% | 19.4% | 36.2% | 8.7% | 17.2% | 51.8% | 60.1% |
| *MiewID-FT* | Cosine similarity | - | 60.2% | 73.8% | 91.3% | 60.2% | 65.0% | 99.0% | 100.0% |
| *MegaDescriptor-L-224* | Cosine similarity | 39.3 | 3.9% | 11.3% | 27.3% | 3.9% | 9.2% | 43.4% | 59.8% |
| *Two-stage pipeline* | | 2,307 | 95.6% | 97.0% | 97.6% | 96.9% | 97.8% | 97.8% | - |

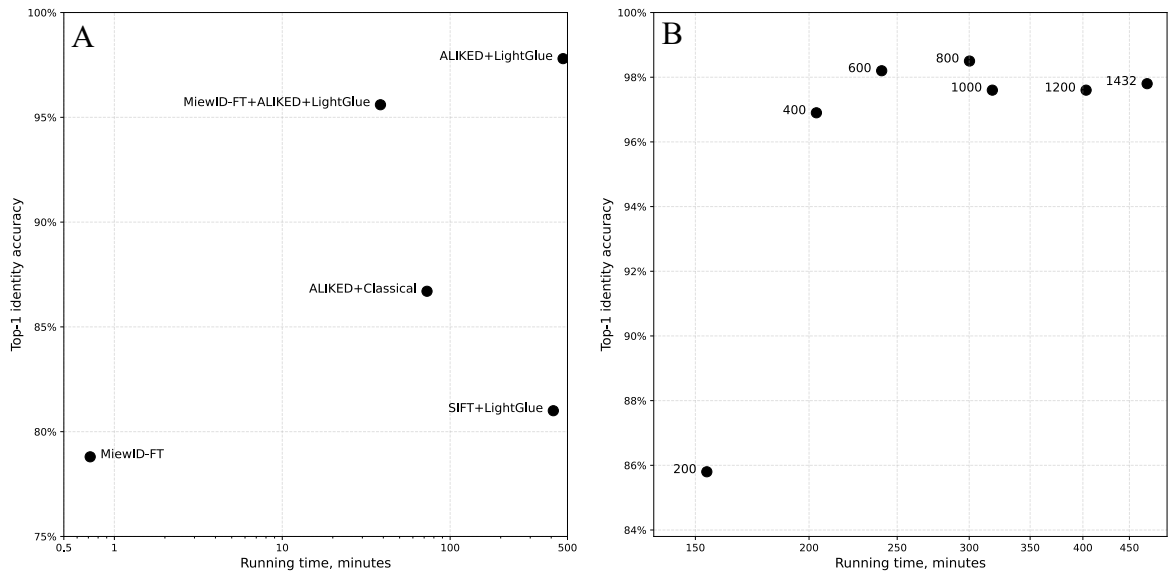

**Figure 3. Accuracy versus running time trade-off.** Top-1 identity accuracy (y-axis) versus running time (x-axis in log scale) for different pipelines evaluated in this study. **(A)** ALIKED+LightGlue is clearly the most accurate, MiewID-FT (finetuned) the fastest (MegaDescriptor models has similar running times), and our proposed two-stage pipeline (MiewID-FT+ALIKED+LightGlue) a good compromise with an order-of-magnitude lower running time than ALIKED+LightGlue alone. **(B)** When varying the maximum number of ALIKED keypoints used by LightGlue from 200 to 1432, top-1 identity accuracy peaks at 800 keypoints while reducing running time by 33% compared to 1432 keypoints, consistent with ranking candidates by the number of matches without an additional spatial consistency check, e.g., RANSAC (Fischler and Bolles 1981). Running time and accuracy were measured on the entire labeled set of 1,233 images, except for MiewID-FT, which was finetuned on the traing set, and therefore only its accuracy was evaluated on the validation set (232 images).

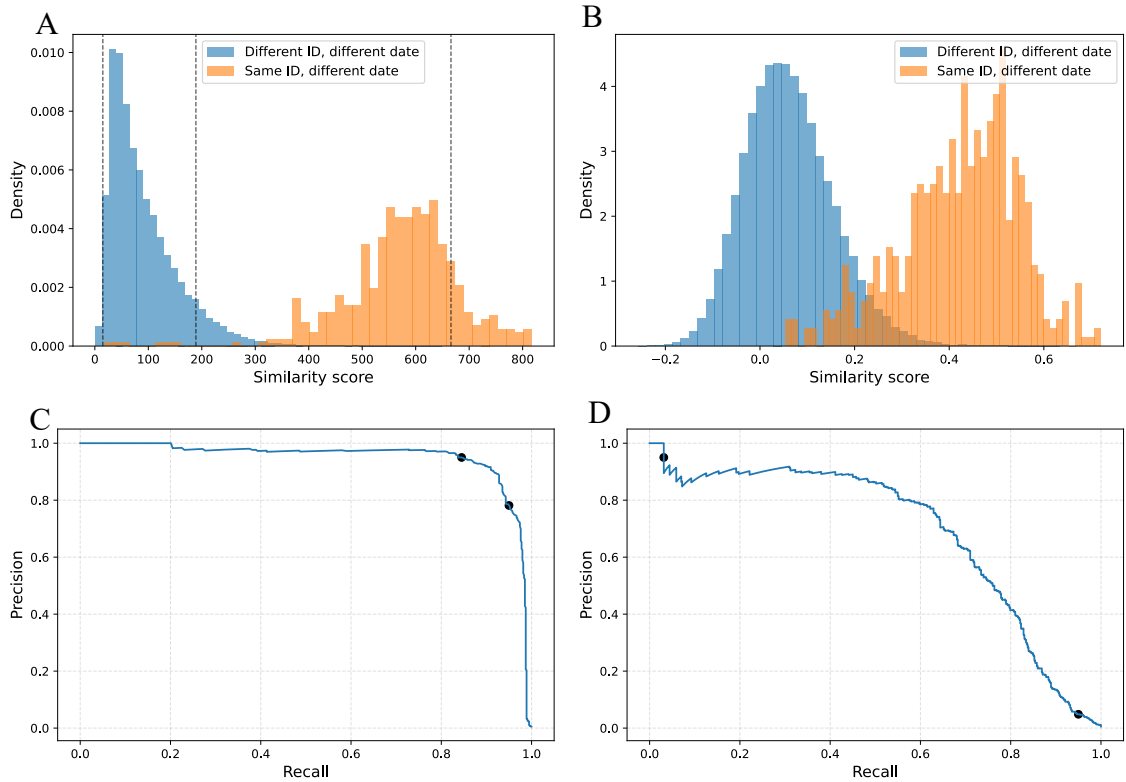

**Figure 4. Open-set matching and precision-recall curve.** Histograms of match similarities between pairs of images of different individuals from different dates (blue; n=697,937 pairs) and pairs of images of the same individual from different days (orange; n=1,629 pairs) for **(A)** ALIKED+LightGlue (dashed lines for the number of matches in Figure 2) and **(B)** finetuned MiewID on the labeled dataset. ALIKED+LightGlue showed a much better separation of similarity scores (despite MiewID having been finetuned on 80% of this dataset). **(C,D)** This separation leads to a superior precision-recall curve (C: ALIKED+LightGlue, D: finetuned MiewID), which shows the precision (y-axis, TP/(TP+FP)) as a function of recall (x-axis, TP/(TP+FN)) as the similarity threshold is varied. Highlighted operating points: at recall=95% with precision=78% at match similarity 384 (C) and with precision=4.8% at cosine similarity 0.18 (D); and at precision=95% with recall=85% at match similarity 479 (C) and with recall=3.1% at cosine similarity 0.633 (D) (Unlike ROC curve, the x-axis here is recall rather than false-positive rate.)

**Discussion**

Our study demonstrates that individual re-identification of the Hula painted frog can be automated with near-perfect accuracy using modern deep local feature matching, even without any training on species-specific images. In a strict zero-shot setting, ALIKED+LightGlue achieved 97.8% top-1 closed-set identification accuracy, whereas two state-of-the-art global features models pretrained on large multi-species datasets performed poorly (≤8.7% top-1, ≤36.2% top-10). Fine-tuning the global embeddings model MiewID on our labeled dataset substantially improved its performance (60.2% top-1, 91.3% top-10), but it still lagged behind the local-feature pipeline and required extensive effort to train and validate. Furthermore, we tested previous-generation local feature matching as a baseline for comparison, but its matching accuracy remained clearly below that of ALIKED+LightGlue (≤87% top-1, ≤94.6% top-10), underscoring the benefit of combining learned local features with a modern deep matching architecture. Together, these results suggest that in our capture-based monitoring setting for Hula painted frogs, which features highly distinctive ventral patterns and non-standardized field imagery, deep local geometric matching provides the most reliable zero-shot solution, while fine-tuned global embeddings can serve as a fast, but imperfect, first-pass retrieval mechanism; how broadly this performance advantage generalizes to other amphibians or imaging protocols remains to be tested.

These findings can be viewed in the context of three "generations" of wildlife re-identification methods: (i) classical approaches using hand-crafted keypoints and local descriptors, e.g., SIFT-based systems such as HotSpotter (Crall et al. 2013) and Wild-ID (Bolger et al. 2012), (ii) deep global embedding models such as MiewID (Otarashvili et al. 2024) and MegaDescriptor (Čermák et al. 2023), and (iii) more recent deep local feature pipelines like ALIKED+LightGlue that combine learned keypoints and descriptors with graph-based deep geometric matching (Zhao et al. 2023; Lindenberger et al. 2023). Classical local-feature pipelines showed that local texture patterns can uniquely identify individuals, but they struggle with strong non-rigid deformations and often require careful user intervention. Deep global-feature models, which now underpin many large-scale platforms, are efficient and powerful when test images are close to the training domain—for example, large terrestrial mammals that dominate current training corpora—but they are also sensitive to domain shift, and performance can degrade substantially for underrepresented taxa such as amphibians. Furthermore, because these models compress all information into a single vector, they are particularly vulnerable to shifts induced by ontogenetic growth and pose-related deformation. By contrast, our results show that modern deep local features models, combined with geometric reasoning over keypoint graphs, can retain high discriminative power even when body shape and spot geometry change markedly between surveys, without relying on species-specific training data.

The performance gap between zero-shot global- and local-feature models in our dataset highlights two limitations of current global-feature models for amphibian monitoring. First, their training corpora are dominated by mammals and marine megafauna, so domain shift to small, moist, high-specular amphibians can be substantial. Second, global-feature models assume that the mapping from appearance to identity can be learned largely from texture and coarse shape, whereas in Latonia the relevant information resides in a deformable, quasi-topological pattern of spots on soft tissue. Even after supervised fine-tuning on our labeled frogs, MiewID did not close the gap to ALIKED+LightGlue, suggesting that the bottleneck is not only the volume of training data but also the inductive bias of global pooling architectures, which are not explicitly designed to be invariant to large, non-uniform geometric transformations.

From an operational perspective, our results support a two-stage architecture that combines the complementary strengths of global- and local-feature models for long-term monitoring. In our implementation, a finetuned global-feature model is used as a fast Stage-1 retrieval mechanism to rank the gallery and produce a short candidate list per query in less than 1 minute, while the computationally expensive local-feature matcher is applied only to the top-k candidates to refine the ranking. This design is consistent with standard practice in large-scale image retrieval and re-identification, where a fast method is used for initial retrieval and a slower method is then used to re-rank a small shortlist (Barbarani et al. 2023). Our benchmarking shows that this hybrid strategy recovers most of the accuracy of full local matching (k=100, top-1 identity accuracy 98.1%, ~38 minutes) while reducing the number of LightGlue evaluations by orders of magnitude (~38 minutes instead of 7.7 hours on the entire labeled dataset), making the system more scalable as the photo gallery grows. The exact choice of k is a tunable engineering parameter that can be adjusted based on available compute and desired latency, rather than a fixed "standard."

The distribution of similarity scores also has important implications for open-set identification and quality control. We observed a strong separation between the match similarity of image pairs belonging to the same individual and that of different individuals, with only a narrow region of overlap (Figure 4). This separation implies that a simple threshold on the number of ALIKED+LightGlue matches can reject most false positives while discarding relatively few true recaptures, enabling open-set operation in which query images of previously unseen individuals can be flagged for assignment of a new identity. In practical monitoring workflows, such a threshold could be used to route uncertain cases to human experts while automatically confirming high-confidence matches, thereby reducing both error rates and manual workload.

Our results have direct consequences for conservation practice. The Hula painted frog has an extremely small and geographically restricted population, and robust capture–recapture estimates are critical for assessing its status and guiding management. Manual photographic matching becomes impractical as image galleries accumulate over many years, particularly when individuals are recaptured across multiple growth stages. By automating individual identification with high accuracy, our pipeline can substantially reduce processing time and observer bias and enable more frequent and larger-scale surveys. Because the method is built entirely from publicly available pretrained models and open-source code, it can be adopted or extended by other monitoring programs with similar imaging protocols and individually identifiable patterns.

At the same time, several limitations of our study should be acknowledged. First, our evaluation is restricted to a single species, site, and imaging protocol: high-resolution ventral photographs of frogs taken at relatively close range under controlled conditions. The excellent performance of the deep local-feature pipeline in this setting does not imply that they will systematically outperform global-feature models for low-resolution, motion-blurred, or heavily occluded camera-trap imagery, where stable keypoints may be sparse or unreliable and global-feature models may retain an advantage. Second, our labeled dataset contains only 191 identities, with relatively few individuals recaptured over many years; although this is typical for critically endangered species, it limits our ability to explore more complex temporal effects such as changes in pattern visibility with age or disease. Third, our ground-truth identities were established via a semi-automated pipeline based on Wild-ID (Bolger et al. 2012) followed by visual confirmation, which may contain a small amount of residual label noise. Finally, although our two-stage system is efficient relative to exhaustive local matching or manual identification, its computational cost still scales with dataset size. This is unlikely to be limiting for critically endangered species, such as the Latonia, but could become a bottleneck for high-throughput monitoring of more abundant taxa or multi-species archives and may require additional engineering for very large databases. Future work should therefore test whether the advantages of deep local geometric matching observed here extend to other deformable taxa (e.g. salamanders, caudates, or small fishes) and to different anatomical views, imaging devices, and environmental conditions. On the global-feature side, architectures that preserve higher spatial resolution or explicitly encode deformation, as well as hybrid schemes that fuse global and local cues at the representation level rather than only at the retrieval stage, may narrow the current performance gap.

To realize its practical potential for conservation decision-making, we have integrated our pipeline for re-identification of Hula painted frogs into a web application with feedback between automatic predictions and expert verification. The application allows researchers to upload query images directly from the field, automatically processes them through our pipeline, and presents ranked candidate matches alongside match scores and the gallery images for each predicted match. The expert user can manually review matches, accept high-confidence identifications, and reject low-confidence identifications. The application can then produce a report of identities and their observed dates for downstream population size estimation using mark-recapture analysis.

Overall, our results provide evidence that for capture-based monitoring of species with strongly deformable body patterns, deep local feature matching with geometric verification should be considered a leading candidate, and often a default, approach for reliable individual re-identification.

**Acknowledgements.** We thank Ofir Levy and Tsur Herman for advice and to Yonatan Ram Simon for technical assistance. This work was supported by Edmond J Safra Center for Bioinformatics at Tel-Aviv University (YR) and the AI and Data Science Center at Tel-Aviv University (YR).

**References**

Barbarani, Giovanni, Mohamad Mostafa, Hajali Bayramov, et al. 2023. "Are Local Features All You Need for Cross-Domain Visual Place Recognition?" arXiv:2304.05887. Preprint, arXiv, April 12. https://doi.org/10.48550/arXiv.2304.05887.

Beery, Sara, Dan Morris, and Siyu Yang. 2019. "Efficient Pipeline for Camera Trap Image Review." arXiv:1907.06772. Preprint, arXiv, July 15. https://doi.org/10.48550/arXiv.1907.06772.

Berton, Gabriele, Kevin Musgrave, and Carlo Masone. 2025. "All You Need to Know About Training Image Retrieval Models." arXiv:2503.13045. Preprint, arXiv, March 17. https://doi.org/10.48550/arXiv.2503.13045.

Bolger, Douglas T., Thomas A. Morrison, Bennet Vance, Derek Lee, and Hany Farid. 2012. "A Computer-Assisted System for Photographic Mark–Recapture Analysis." *Methods in Ecology and Evolution* 3 (5): 813–22. https://doi.org/10.1111/j.2041-210X.2012.00212.x.

Bradski, G. 2000. "The OpenCV Library." *Dr. Dobb's Journal of Software Tools*.

Čermák, Vojtěch, Lukas Picek, Lukáš Adam, Lukáš Neumann, and Jiří Matas. 2024. "WildFusion: Individual Animal Identification with Calibrated Similarity Fusion." Version 1. Preprint, arXiv. https://doi.org/10.48550/ARXIV.2408.12934.

Čermák, Vojtěch, Lukas Picek, Lukáš Adam, and Kostas Papafitsoros. 2023. "WildlifeDatasets: An Open-Source Toolkit for Animal Re-Identification." arXiv:2311.09118. Preprint, December 14. https://doi.org/10.48550/arXiv.2311.09118.

Crall, Jonathan P., Charles V. Stewart, Tanya Y. Berger-Wolf, Daniel I. Rubenstein, and Siva R. Sundaresan. 2013. "HotSpotter — Patterned Species Instance Recognition." *2013 IEEE Workshop on Applications of Computer Vision (WACV)*, January, 230–37. https://doi.org/10.1109/WACV.2013.6475023.

Deng, Jiankang, Jia Guo, Tongliang Liu, Mingming Gong, and Stefanos Zafeiriou. 2020. "Sub-Center ArcFace: Boosting Face Recognition by Large-Scale Noisy Web Faces." In *Computer Vision – ECCV 2020*, edited by Andrea Vedaldi, Horst Bischof, Thomas Brox, and Jan-Michael Frahm, vol. 12356. Lecture Notes in Computer Science. Springer International Publishing. https://doi.org/10.1007/978-3-030-58621-8_43.

Deng, Jiankang, Jia Guo, Jing Yang, Niannan Xue, Irene Kotsia, and Stefanos Zafeiriou. 2022. "ArcFace: Additive Angular Margin Loss for Deep Face Recognition." *IEEE Transactions on Pattern Analysis and Machine Intelligence* 44 (10): 5962–79. https://doi.org/10.1109/TPAMI.2021.3087709.

Ferner, John William. 2007. *A Review of Marking and Individual Recognition Techniques for Amphibians and Reptiles*. Society for the Study of Amphibians and Reptiles Salt Lake City, Utah.

Fischler, Martin A, and Robert C Bolles. 1981. "Random Sample Consensus: A Paradigm for Model Fitting with Apphcatlons to Image Analysis and Automated Cartography." *Communications of the ACM* 24 (6).

Kirillov, Alexander, Eric Mintun, Nikhila Ravi, et al. 2023. "Segment Anything." *Proceedings of the IEEE/CVF International Conference on Computer Vision (ICCV)*, October, 4015–26.

Lindenberger, Philipp, Paul-Edouard Sarlin, and Marc Pollefeys. 2023. "LightGlue: Local Feature Matching at Light Speed." arXiv:2306.13643. Preprint, June 23. https://doi.org/10.48550/arXiv.2306.13643.

Liu, Ze, Yutong Lin, Yue Cao, et al. 2021. "Swin Transformer: Hierarchical Vision Transformer Using Shifted Windows." arXiv:2103.14030. Preprint, arXiv, August 17. https://doi.org/10.48550/arXiv.2103.14030.

Lorm, Tijmen A. de, Catharine Horswill, Daniella Rabaiotti, et al. 2023. "Optimizing the Automated Recognition of Individual Animals to Support Population Monitoring." *Ecology and Evolution* 13 (7): e10260. https://doi.org/10.1002/ece3.10260.

Loshchilov, Ilya, and Frank Hutter. 2019. "Decoupled Weight Decay Regularization." arXiv:1711.05101. Preprint, arXiv, January 4. https://doi.org/10.48550/arXiv.1711.05101.

Lowe, David G. 2004. "Distinctive Image Features from Scale-Invariant Keypoints." *International Journal of Computer Vision* 60 (2): 91–110. https://doi.org/10.1023/B:VISI.0000029664.99615.94.

Luedtke, Jennifer A., Janice Chanson, Kelsey Neam, et al. 2023. "Ongoing Declines for the World's Amphibians in the Face of Emerging Threats." *Nature* 622 (7982): 308–14. https://doi.org/10.1038/s41586-023-06578-4.

Matthé, Maximilian, Marco Sannolo, Kristopher Winiarski, et al. 2017. "Comparison of Photo-matching Algorithms Commonly Used for Photographic Capture–Recapture Studies." *Ecology and Evolution* 7 (15): 5861–72. https://doi.org/10.1002/ece3.3140.

*Measuring and Monitoring Biological Diversity: Standard Methods for Amphibians*. 1994. https://pubs.usgs.gov/publication/5200175.

Musgrave, Kevin, Serge Belongie, and Ser-Nam Lim. 2020. "A Metric Learning Reality Check." arXiv:2003.08505. Preprint, September 16. https://doi.org/10.48550/arXiv.2003.08505.

Nemesházi, Edina, Nikolett Ujhegyi, Zsanett Mikó, Andrea Kásler, Vera Lente, and Veronika Bókony. 2025. "Photo-Based Individual Identification Is More Reliable than Visible Implant Elastomer Tags or Toe-Tipping in Young Agile Frogs." Preprint, Ecology, June 25. https://doi.org/10.1101/2025.06.19.660530.

Nepovinnykh, Ekaterina, Ilia Chelak, Tuomas Eerola, et al. 2024. "Species-Agnostic Patterned Animal Re-Identification by Aggregating Deep Local Features." *International Journal of Computer Vision* 132 (9): 4003–18. https://doi.org/10.1007/s11263-024-02071-1.

Otarashvili, Lasha, Tamilselvan Subramanian, Jason Holmberg, J. J. Levenson, and Charles V. Stewart. 2024. "Multispecies Animal Re-ID Using a Large Community-Curated Dataset." arXiv:2412.05602. Preprint, arXiv, December 7. https://doi.org/10.48550/arXiv.2412.05602.

Perl, R. G. Bina, Sarig Gafny, Eli Geffen, Yoram Malka, Sharon Renan, and Miguel Vences. 2018. "Notes on Post-Metamorphic Colour Pattern Changes in the Hula Painted Frog (Latonia Nigriventer): How Realistic Is Re-Identification of Juveniles?" *Herpetology Notes* 11 (May): 475–80.

Perl, R. G. Bina, Sarig Gafny, Yoram Malka, et al. 2017. "Natural History and Conservation of the Rediscovered Hula Painted Frog, Latonia Nigriventer." *Contributions to Zoology* 86 (1): 11–37. https://doi.org/10.1163/18759866-08601002.

Perl, R. G. Bina, Eli Geffen, Yoram Malka, et al. 2018. "Population Genetic Analysis of the Recently Rediscovered Hula Painted Frog (Latonia Nigriventer) Reveals High Genetic Diversity and Low Inbreeding." *Scientific Reports* 8 (1): 5588. https://doi.org/10.1038/s41598-018-23587-w.

Tan, Mingxing, and Quoc V. Le. 2021. "EfficientNetV2: Smaller Models and Faster Training." arXiv:2104.00298. Preprint, arXiv, June 23. https://doi.org/10.48550/arXiv.2104.00298.

Zhao, Xiaoming, Xingming Wu, Weihai Chen, Peter C. Y. Chen, Qingsong Xu, and Zhengguo Li. 2023. "ALIKED: A Lighter Keypoint and Descriptor Extraction Network via Deformable Transformation." arXiv:2304.03608. Preprint, April 16. https://doi.org/10.48550/arXiv.2304.03608.